\pdfoutput=1

\documentclass[11pt]{article}

\usepackage{EMNLP2023}
\usepackage{times}
\usepackage{latexsym}
\usepackage{comment}
\usepackage[T1]{fontenc}
\usepackage[utf8]{inputenc}
\usepackage{microtype}
\usepackage{inconsolata}

\usepackage{multirow}

\usepackage{graphicx} 
\usepackage{diagbox} 
\usepackage{makecell}
\usepackage{stfloats}

\title{Unnatural Error Correction: GPT-4 Can Almost Perfectly Handle Unnatural Scrambled Text}

\author{Qi Cao, Takeshi Kojima, Yutaka Matsuo, Yusuke Iwasawa\\
         The University of Tokyo, Japan\\
         \texttt{\{qi.cao,t.kojima,matsuo,iwasawa\}@weblab.t.u-tokyo.ac.jp} \\}

\begin{document}
\maketitle
\begin{abstract}
While Large Language Models (LLMs) have achieved remarkable performance in many tasks, much about their inner workings remains unclear.
In this study, we present novel experimental insights into the resilience of LLMs, particularly GPT-4, when subjected to extensive character-level permutations.
To investigate this, we first propose the \emph{Scrambled Bench}, a suite designed to measure the capacity of LLMs to handle scrambled input, in terms of both recovering scrambled sentences and answering questions given scrambled context. The experimental results indicate that most powerful LLMs demonstrate the capability akin to \emph{typoglycemia}\footnote{Typoglycemia is an internet meme without specific scientific backing. Nonetheless, people do exhibit the ability to recognize scrambled common words to some extent. Our focus is not on examining this ability of humans, we use the concept for illustrative purposes.}, a phenomenon where humans can understand the meaning of words even when the letters within those words are scrambled, as long as the first and last letters remain in place.
More surprisingly, we found that only GPT-4 nearly flawlessly processes inputs with \emph{unnatural errors}, even under the extreme condition, a task that poses significant challenges for other LLMs and often even for humans. Specifically, 
GPT-4 can almost perfectly reconstruct the original sentences from scrambled ones, decreasing the edit distance by 95\%, even when all letters within each word are entirely scrambled. It is counter-intuitive that LLMs can exhibit such resilience despite severe disruption to input tokenization caused by scrambled text.\footnote{Code will be available at \url{https://github.com/ccqq77/unnatural-error-correction}.}

\end{abstract}

\section{Introduction}

Large language models (LLMs) demonstrate impressive proficiency across a range of tasks, with certain capabilities emerging as the models scale up in size---a phenomenon commonly known as \textit{emergent abilities}.~\citep{wei2022emergent}. 
As LLMs become more ``intelligent'', many new benchmarks have been proposed~\citep{liang2022holistic, qin2023chatgpt} to investigate the abilities of LLMs. Nonetheless, the internal mechanisms underlying the abilities of LLMs remain enigmatic.

Several studies investigate the behavior of LLMs given some input perturbations. 
For example,~\citet{sinha-etal-2021-masked, sinha-etal-2021-unnatural, abdou-etal-2022-word} investigate the influence of \emph{word-level} permutations and discuss the phenomenon that models are insensitive to permutations of word order which corrupt the original syntax, in some downstream tasks (e.g., natural language inference).
These results are particularly interesting because they challenge the common assumption of the inner workings of LLMs, i.e., LLMs understand human-like syntax to some extent and use it to understand sentences. 

\begin{figure}[t]
	
	\begin{minipage}{1.0\hsize}
		\centering
        \captionsetup{width=.95\linewidth}
        \includegraphics[width=0.97\linewidth]{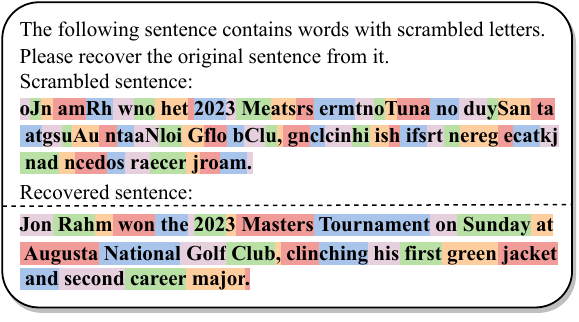}
        \vspace{-0.1cm}
	\end{minipage} \\
	
	\begin{minipage}{1.0\hsize}
		\centering
        \captionsetup{width=.95\linewidth}
	    \includegraphics[width=0.97\linewidth]{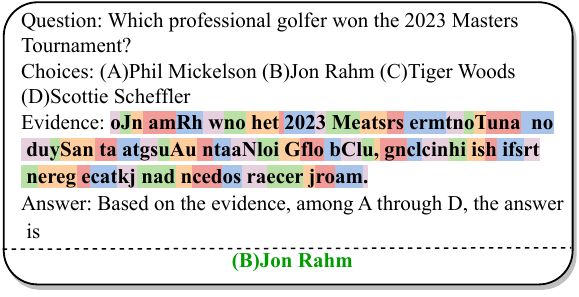}
        \vspace{-0.2cm}
    	\caption{GPT-4 can recover original sentence from scrambled sentence (upper) and answer question based on scrambled context (lower), even if the tokenization drastically changes. The colors indicate the division of sub-words during the tokenization.}
     \label{fig:gpt4_token}
     \vspace{-0.35cm}
	\end{minipage}
\end{figure}

Going beyond the effect of word-level permutations, we investigate the ability of LLMs under \emph{character-level} permutations. LLMs are supposed to rely on the tokenizers to turn natural language into the form that LLMs can perceive. It would be counter-intuitive if LLMs could effectively handle text containing \emph{unnatural} permutations that significantly alter tokenization. In other words, we propose the following research question: 
\emph{Deos the oredr of ltteers in wrods mttaer for LLMs?}
Note that the above sentence contains scrambled words, but humans can somehow recognize and understand such a sentence, as several cognitive studies have explored~\citep{rawlinson2007significance, mason1982recognition, johnson2012importance}. 
Analyzing the robustness of LLMs against such character-level permutations can shed light on their word comprehension capabilities and reveal differences between various LLMs and human understanding.

To this end, this paper first constructs \emph{Scrambled Bench}, which converts existing benchmarks into a test suite to measure the ability of LLMs to handle scrambled text. 
We designed two types of tasks: (1) \emph{Scrambled Sentence Recovery}, which tests the capability of LLMs to reconstruct original sentences from scrambled ones, and (2) \emph{Scrambled Question Answering}, which measures how well LLMs can answer questions when some context is scrambled.
Note that since the slight change in letter-order within a word drastically changes the tokenized output (see~\autoref{fig:gpt4_token})
, it is questionable whether LLMs can recognize the scrambled words in a sentence. 
Counter-intuitively, we show that most powerful LLMs are able to handle scrambled sentences to varying degrees, when we scramble words while keeping the first and last letters unchanged.
More surprisingly, we found that only GPT-4 can almost flawlessly process inputs with \emph{unnatural errors}, even under extreme conditions.
That is, even when we scramble all letters in words, GPT-4 manages to handle such input --- a significantly challenging task for other models and even humans. For instance, GPT-4 can reconstruct original sentences to near-perfect recovery rate in the extreme scenario, as in~\autoref{fig:gpt4_token}. 
\begin{figure*}[t]
    \centering
    \captionsetup{width=.95\linewidth}
    \includegraphics[width=.96\linewidth]{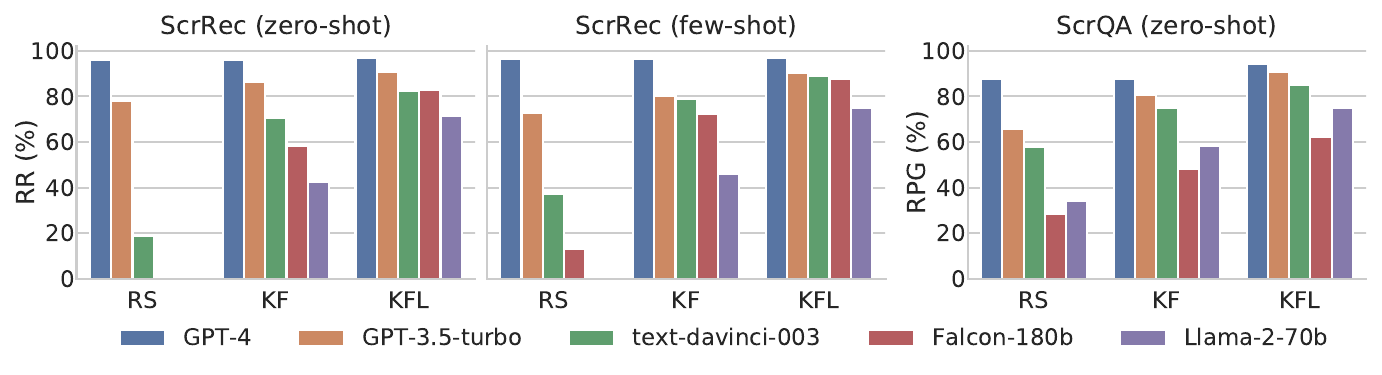}
    \vspace{-0.47cm}
    \caption{Experimental results using scrambled RealtimeQA dataset, including performance on ScrRec in a zero-shot setting (left) and a few-shot setting (middle) and performance on ScrQA in a zero-shot setting (right). The missing bars represent values less than zero.}
    \label{fig:sr_and_sqa_result}
    \vspace{-0.3cm}
\end{figure*}

\section{Related Works}
The most related works are the studies investigating the effects of word or sub-word level perturbations~\cite{sinha-etal-2021-masked, sinha-etal-2021-unnatural, pham-etal-2021-order, abdou-etal-2022-word} and the studies evaluating the robustness of LLMs~\cite{wang2023robustness, zhu2023promptbench}. To the best of our knowledge, no existing studies have investigated LLMs' ability to handle character-level permutations, particularly those of an extremely high level that drastically change tokenization. Our study aims to fill this gap.~\autoref{tab: related_work} in~\autoref{sec:app_related_work} categorizes the prior studies and demonstrates the position of our study.

For evaluations, we assess the ability to reconstruct the original text from its scrambled version and to answer questions using the scrambled text as a reference.
It differs with typographical error correction~\citep{shah-de-melo-2020-correcting,sun-etal-2022-error}, as (i) we do not train models to correct errors, i.e., we measure the ability of LLMs, and (ii) we add much more severe noises than natural typographical errors.
The word unscrambling task in BigBench~\citep{srivastava2023beyond} is similar to our recovery task. However, it is more akin to a wordplay puzzle rather than a task for comprehending scrambled text, since it includes only single common words and no context is given.

\section{Scrambled Bench}
We propose two tasks (as in~\autoref{fig:gpt4_token}) to evaluate the capability of LLMs to handle scrambled text:
\paragraph{(1) Scrambled Sentence Recovery (ScrRec).}
In this task, we provide a sentence containing scrambled words to LLMs and then ask them to recover original sentence from it. This task can be utilized to directly measure the capability of LLMs to recognize and reconstruct the scrambled words in a sentence. 

\paragraph{(2) Scrambled Question Answering (ScrQA).} 

While ScrRec can directly measure the capability to comprehend and process scrambled text, it is an ``unusual'' task for LLMs. If a model does not perform well on the recovery task, there are two possible reasons: (i) having difficulty following the instructions and (ii) not being able to recover sentences. To distinguish them, we measure the ability to accomplish a standard task (i.e., QA) given scrambled context. Specifically, we scramble the context that contains essential information for answering questions and then assess the models based on the variations in their performances.
\subsection{Dataset Creation}
Data contamination is a significant concern when assessing LLMs, when the test data for downstream tasks has been inadvertently included in the training set. This issue is critical for our tasks, as the reliability of the results could be compromised if some models simply memorized original content.
To mitigate this, we primarily utilize a scrambled version of RealtimeQA~\citep{kasai2022realtime} for evaluations in this study. 
RealtimeQA is a dynamic question answering dataset that weekly announces questions about recent news that are unlikely to be memorized by the current LLMs.
Specifically, we collect the most recent data (2023/03/17--2023/08/04) from RealtimeQA (totally 419 samples) and process the evidence sentences to construct samples for ScrRec and ScrQA. Finally, 418 samples are selected for ScrRec (removing a duplicate sentence), and 346 samples are selected for ScrQA (manually eliminating 73 samples when the provided evidence does not provide sufficient information to answer the corresponding question).

Besides, we also introduce two additional datasets: DREAM~\cite{sun-etal-2019-dream} and AQuA-RAT~\citep{ling2017program}. DREAM is a dialogue-based multiple-choice reading comprehension dataset. AQuA-RAT is a dataset of math word problems necessitating multi-step reasoning for their resolution. For DREAM dataset, we constructed the dataset by selecting 1025 samples with annotated categories from the development and test sets and then scrambling the dialogue part of each question. For AQuA-RAT dataset, we adopt the few-shot Chain of Thought (CoT) setting as in~\citealt{wei2022chain} and evaluate LLMs with scrambled questions in samples and demonstrations.

For each dataset, we generate scrambled text with various scramble types and rates. \textbf{(1) Randomly Scramble (RS)}. For each sentence, we randomly select a certain rate (20\%, 50\%, 100\% in our experiments\footnote{In the latter part of the paper, RS denotes 100\% randomly scrambling, unless a different percentage is specified.}) of words and randomly scramble all letters in each selected word (Arabic numerals are kept invariant). \textbf{(2) Keep First (KF)}. We keep the first letter in each word unchanged and randomly scramble the letters in other positions. \textbf{(3) Keep First and Last (KFL)}. We keep the first and last letter in each word unchanged and randomly scramble the letters in other positions. We use KF and KFL to investigate the effects of different scramble types in word recognition for LLMs.

\subsection{Metrics}
For ScrRec, the average Edit Distance (ED)~\citep{levenshtein1966binary} between original sentences and recovered sentences is a natural metric to quantify the performance. Besides, we define Recovery Rate (RR) to measure the proportion of ED reduced in recovered sentences, which makes the performance comparison in different settings more straightforward as:
\begin{equation}
    RR = \frac{\sum_i^n ED({ori}_i, {rec}_i)-\sum_i^n ED({ori}_i, {scr}_i)}{\sum_i^n ED({ori}_i, {scr}_i)}
\end{equation}
where $ori_i$, $scr_i$, $rec_i$ mean original sentence, scrambled sentence, and recovered sentence of sample $i$, respectively. If a model is capable of recovering scrambled text to some extent, its RR should range between 0 and 100\%.

For ScrQA, accuracy is a natural metric to measure the performance. But varying capabilities of models on original questions make it hard to compare the performance among models. So, Relative Performance Gain (RPG) is defined to mitigate the differences and make evaluations focus on how well models are able to comprehend scrambled text in comparison to original text as:
\begin{equation}
    RPG = \frac{Acc_{scr} - Acc_{sub}}{Acc_{ori} - Acc_{sub}}
\end{equation}
where $Acc_{ori}$, $Acc_{scr}$, $Acc_{sub}$ are defined as accuracy when using original evidences, scrambled evidences, substituted evidences (in which all letters of words that can be scrambled have been substituted with random letters), respectively. $Acc_{ori}$, $Acc_{sub}$ are served as the higher bound and the lower bound of accuracy, respectively.
Normally, RPG ranges between 0 and 100\%.
\begin{figure*}[t]
    \centering
    \captionsetup{width=.95\linewidth}
    \includegraphics[width=.95\linewidth]{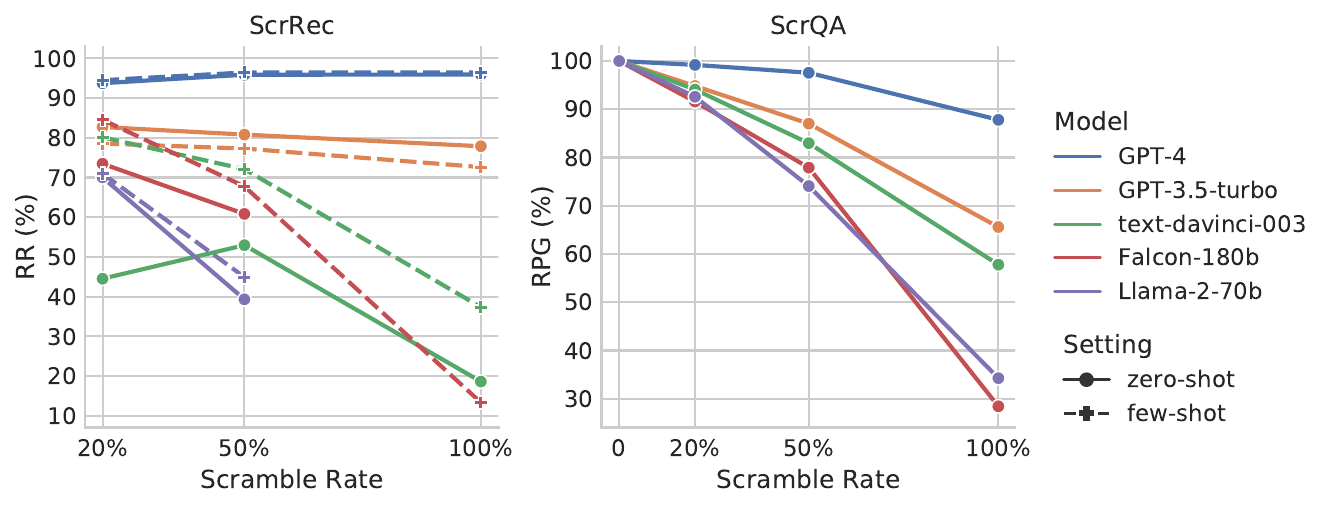}
    \vspace{-0.4cm}
    \caption{The effect of scramble rates on the performance on ScrRec (left) and ScrQA (right) using RealtimeQA dataset. The missing points represent values less than zero.}
    \label{fig:realtimeqa_rate}
    \vspace{-0.2cm}
\end{figure*}
\section{Experiments}

\paragraph{Models and Settings.}
In the experiments, we evaluate the most powerful closed-source LLMs, including text-davinci-003~\cite{brown2020language}, GPT-3.5-turbo and GPT-4~\cite{openai2023gpt4} and the open-source models from Falcon series~\cite{refinedweb}, Llama-2 series~\cite{touvron2023llama}, MPT series~\cite{MosaicML2023Introducing}, UL2 series~\cite{tay2022ul2}, T5 series~\cite{raffel2020exploring, https://doi.org/10.48550/arxiv.2210.11416, xue-etal-2022-byt5}. For the scrambled RealtimeQA dataset, we adopt a zero-shot setting and a few-shot setting with 3-shot exemplars from the wikiQA dataset~\citep{yang2015wikiqa} for ScrRec, while we only conduct experiments in a zero-shot setting for ScrQA (since the task is rather straightforward). For the scrambled DREAM dataset, the setting is also zero-shot ScrQA. For the scrambled AQuA dataset, we adopt a few-shot CoT setting with scrambled demonstrations (in the question part). Note that we are showcasing the results of the top five most proficient LLMs (i.e., GPT-4, GPT-3.5-turbo, text-davinci-003, Falcon-180b, Llama-2-70b) in this section, while full results can be found in~\autoref{sec:full_results} and examples of different tasks can be found in~\autoref{sec:examples}.

\paragraph{Results 1: Effect of different scramble types.}
\autoref{fig:sr_and_sqa_result} show the results on zero-shot ScrRec, few-shot ScrRec, and zero-shot ScrQA, with three scramble types: randomly scramble (RS), keep first (KF), and keep first and last (KFL).
The results show the performance gaps between models are not large in KFL setup. However, except for GPT-4, performance significantly decreases as the difficulty of scramble types increases (KFL, KF, and RS in order).
In contrast, the performance of GPT-4 remains constantly high regardless of the scramble types. 
On ScrRec, RR of GPT-4 is constantly above 95\% for all setups.
On ScrQA, GPT-4 consistently outperforms other models, maintaining a high level of accuracy even as the difficulty of scramble types increases.

\begin{figure*}[t]
    \centering
    \captionsetup{width=.95\linewidth}
    \includegraphics[width=.95\linewidth]{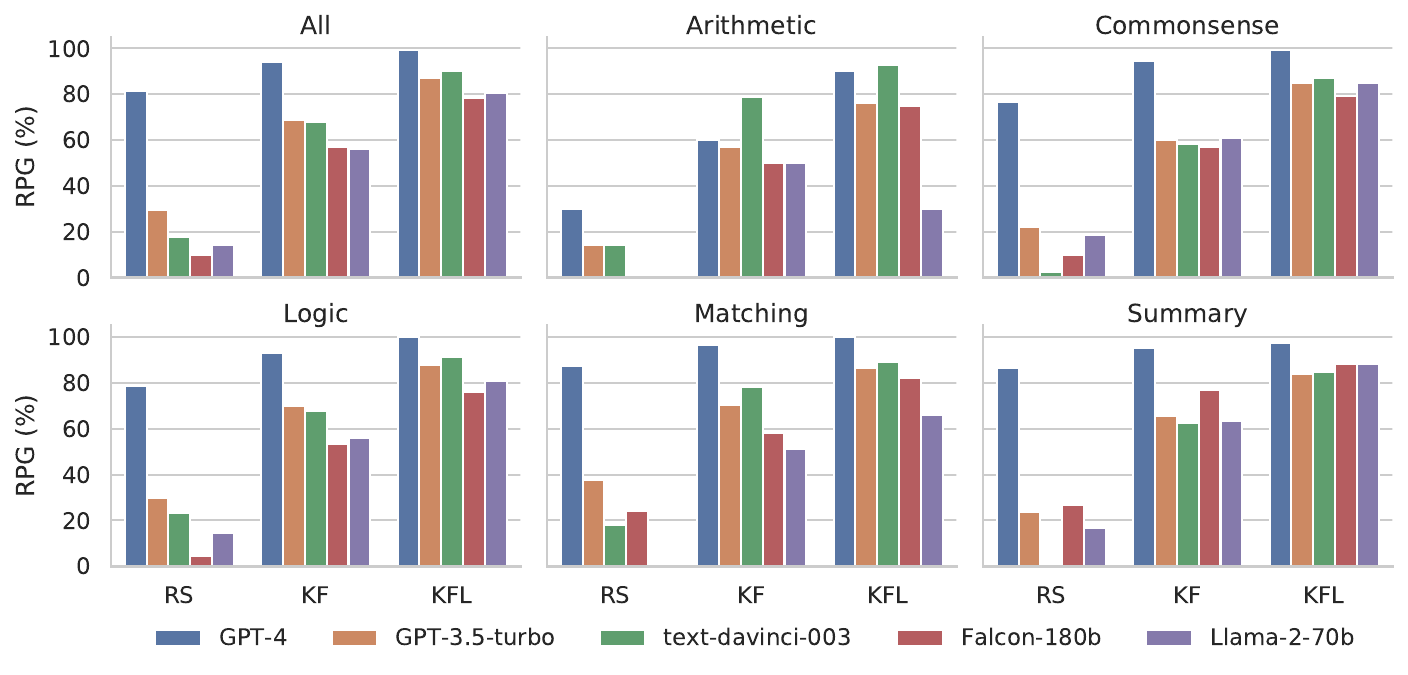}
    \vspace{-0.45cm}
    \caption{Experimental results of ScrQA on scrambled DREAM dataset (zero-shot). Overall performance and performance on different categories of questions. The missing bars represent values less than or equal to zero.}
    \label{fig:dream_result}
    \vspace{-0.114cm}
\end{figure*}

\paragraph{Results 2: Effect of different scramble rates.}
\autoref{fig:realtimeqa_rate} (left) illustrates the relationship between the scramble rates (i.e., the percentages of randomly scrambled words in text) and the performance on ScrRec using scrambled RealtimeQA.
As the scramble rates increases, RR decreases for text-davinci-003, Falcon-180b and Llama-2-70b. 
RR of GPT-3.5-turbo and GPT-4 does not change significantly. GPT-4 outperforms other models by a wide margin, with higher than 95\% RR for most setups (except for 20\% scramble rate).

\autoref{fig:realtimeqa_rate} (right) displays a plot of RPG versus scramble rates for different models on ScrQA using scrambled RealtimeQA. RPG of all models drops as scramble percentage increases. But GPT-4 maintains 87.8\% of its original performance even with 100\% scrambled evidences. The gap between the performance of different models becomes larger when the scramble rates increases.~\autoref{tab:realtime_rec_few},~\autoref{tab:realtime_rec_zero} and~\autoref{tab:realtime_qa} in~\autoref{sec:full_results} show full results on scrambled RealtimeQA, with additional evaluation metrics (ED for ScrRec and Acc for ScrQA).

\begin{table*}[t]
\small
\centering
\captionsetup{width=.95\linewidth}
\scalebox{0.97}{
\begin{tabular}{c|cccc}
\hline
  \multirow{3}{*}{\backslashbox{\textbf{Question}}{\textbf{Exemplars}}}  & \multirow{3}{*}{\textbf{Original}}          & \multirow{3}{*}{\textbf{20\%} }             & \multirow{3}{*}{\textbf{50\%} }             & \multirow{3}{*}{\textbf{100\%}} \\ \\ \\ \hline
  \textbf{Original}             & 66.93 | 52.76 | 46.85 & 70.87 | 51.97 | 47.24 & 72.05 | 54.72 | 45.67 & 70.08 | 54.72 | 46.46 \\
  \textbf{20\%}                 & 66.54 | 49.61 | 40.94 & 74.02 | 51.57 | 40.16 & 68.50 | 52.36 | 42.52 & 71.26 | 50.79 | 45.28 \\ 
  \textbf{50\%}                 & 67.72 | 40.94 | 40.94 & 65.75 | 44.88 | 41.34 & 64.96 | 47.64 | 40.16 & 67.32 | 42.91 | 44.09 \\
  \textbf{100\%}                & 64.17 | 29.92 | 26.77 & 62.60 | 31.89 | 27.17 & 64.96 | 27.17 | 33.47 & 60.24 | 35.04 | 28.35 \\ 
\hline
\end{tabular}}
\vspace{-0.15cm}
\caption{
Experimental results of ScrQA on scrambled AQuA-RAT dataset in a few-shot CoT setting. LLMs are assessed in various scenarios by adjusting scramble rate of questions in examplars and the main question. (left: GPT-4; middle: GPT-3.5-turbo; right: text-davinci-003). Metric: Acc (\%).
}
\label{scrambled_qa_results_aqua}
\vspace{-0.1cm}
\end{table*}
\paragraph{Results 3: Results of additional datasets.}
We test the generality of the finding across datasets by two additional datasets for ScrQA. 
On scrambled DREAM dataset, we evaluate performance not only overall but also on different categories of questions using the annotations (in \autoref{fig:dream_result}). The performance disparities between GPT-4 and other models are more pronounced than those observed on RealtimeQA, possibly since DREAM requires higher-level comprehension of longer texts.
The performance on arithmetic questions tends to be more susceptible to scrambled text compared to other categories, even for GPT-4.
\autoref{scrambled_qa_results_aqua} demonstrates experimental results in a 4-shot CoT setting on scrambled AQuA-RAT dataset (we only assess the three closed-source models here because even original questions in AQuA-RAT are too challenging for most open-source models). We evaluate models with several different scenarios by changing scramble rates in both main question and questions in exemplars. The results show the influence of scrambled exemplars is relatively small. The performance of GPT-3.5-turbo and text-davinci-003 declines markedly as the scramble rate of the main questions reaches 100\%, whereas GPT-4 retains much of its initial performance.~\autoref{tab:dream_qa_all},~\autoref{tab:dream_qa_cat_acc} and~\autoref{tab:dream_qa_cat_rpg} in~\autoref{sec:full_results} show full results on scrambled DREAM dataset.

We also conduct a deeper analysis, the details of which can be found in~\autoref{sec:analysis}.

\section{Conclusion}
In this study, we propose \emph{Scrambled Bench}, a test suite to measure the ability of LLMs to handle scrambled text, including two tasks (i.e., scrambled sentence recovery and scrambled question answering) and construct scrambled datasets based on RealtimeQA, DREAM and AQuA-RAT. 
Despite the scrambled text drastically changes the tokenization, we demonstrate that most powerful LLMs are capable of processing scrambled text to varying degrees. However, most LLMs have difficulty handling text that is scrambled to an extreme degree (i.e., 100\% randomly scrambling).
Surprisingly, for both tasks, GPT-4 shows good results and outperforms other models by a large margin. For the scrambled sentence recovery task, GPT-4 can recover sentences by 95\% edit distance reduction even in 100\% randomly scrambling settings. For the scrambled question answering task, GPT-4 can maintain a very high proportion of its original accuracy using scrambled context.

\section*{Limitations}
For LLMs, there are various ways to disrupt the tokenization of words (e.g., inserting letters, substituting letters). In this study, we only investigate the influence of scrambling the letter-order in words. Investigating the performance of LLMs to handle other situations would be an interesting topic. 

In addition, we have conducted our experiments using only three datasets, RealtimeQA, DREAM and AQuA-RAT. Experiments on more diverse datasets could be another future work. 
Note that the two tasks can be applicable for diverse datasets, and it is easy to extend the analysis.

We investigate the capability of different LLMs to handle scrambled text in different settings. However, it is difficult to conclude the reason why some LLMs are capable of handling these tasks. Especially, the reason why GPT-4 can perform almost perfectly would be an interesting topic worth further investigation. 
We can not access the closed-source models directly and are aware of little information about them (even the exact parameter size of GPT-4). 
This makes investigating the reason difficult.
One potential hypothesis is that this capability might stem from training methods, such as including denoising tasks or utilizing a huge corpus of text data with various errors.
Another one is that this capability could emerges as LLMs scale.
Validating the potential hypotheses could enhance our understanding of the inner workings of LLMs.
\bibliography{anthology,custom}
\bibliographystyle{acl_natbib}

\clearpage
\appendix

\section{Analysis}
\label{sec:analysis}

\subsection{Relationship to Parameter Scale}
\begin{figure*}[!hb]
    \centering
    \captionsetup{width=.95\linewidth}
    \includegraphics[width=.89\linewidth]{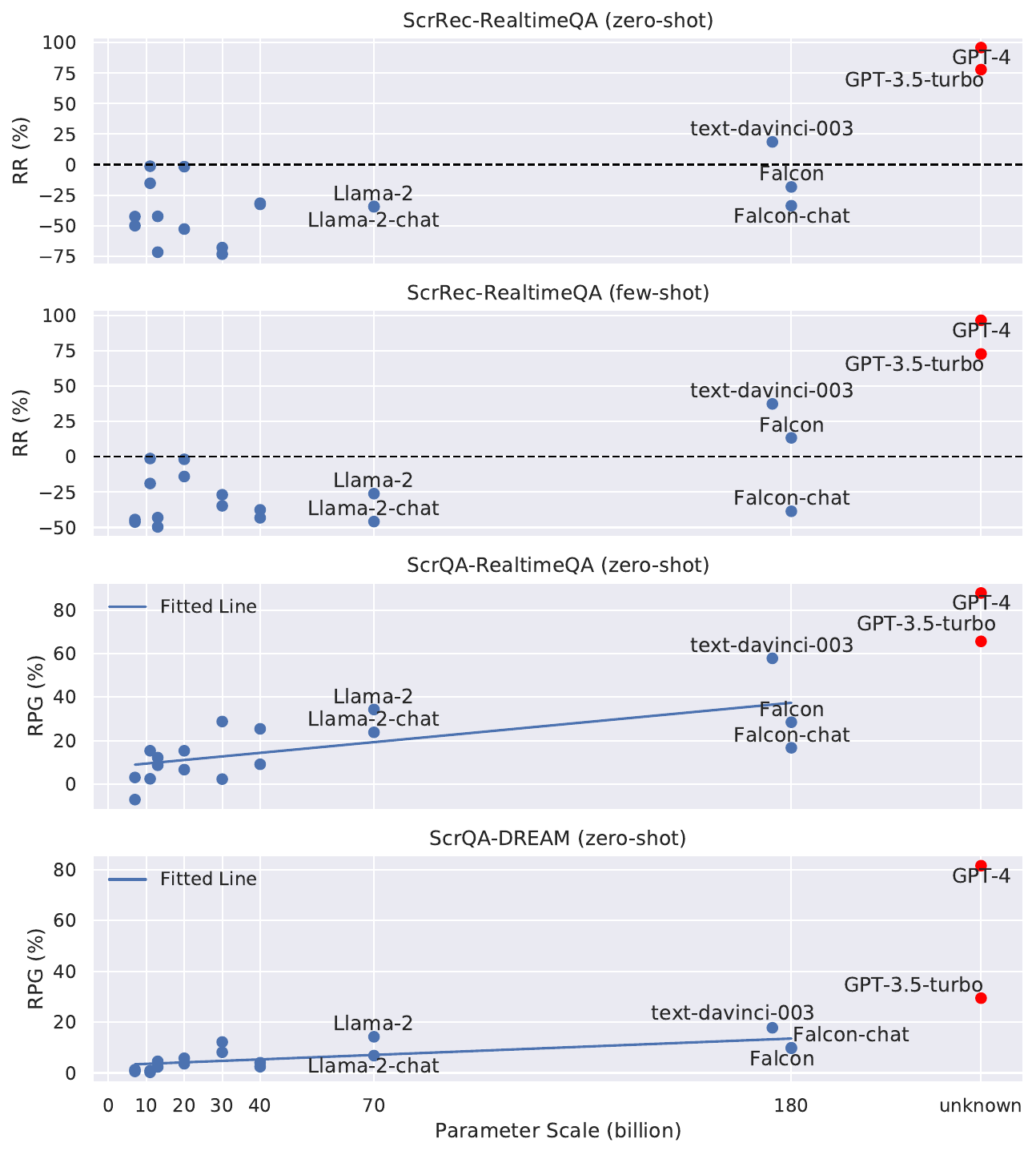}
    \vspace{-0.2cm}
    \caption{Performance of LLMs in relation to their parameter sizes on four tasks, tested on 100\% randomly scrambled text.}
    \label{fig:scale}
    \vspace{-0.cm}
\end{figure*}
By scaling up their parameters, LLMs can significantly enhance their performance across a range of tasks and even acquire emergent abilities that were previously unattainable with smaller parameter sizes.~\citep{wei2022emergent}. 

We investigate the relationship between the ability to handle scrambled text and parameter scale.~\autoref{fig:scale} displays the performance of LLMs in relation to their parameter sizes on 100\% randomly scrambled text to illustrates the relationship (though the parameter sizes of GPT-4 and GPT-3.5-turbo are unknown). On ScrRec-RealtimeQA, only the three closed-source models can achieve a positive recovery rate in the zero-shot setting, while an additional open-source model, Falcon-180b, can achieve a positive recovery rate in the few-shot setting alongside the closed-source models. On ScrQA, performance improves with parameter scaling when using RealtimeQA dataset. However, even when parameters are scaled up to 180 billion, performance improvement on the DREAM dataset is limited due to its longer texts and more diverse questions, which present greater challenges. The top three models in all evaluations are the three closed-source models. Especially, GPT-4 significantly outperforms the others, although its parameter size is unknown.

\subsection{Letter Importance}
\begin{figure*}[ht!]
    \centering
    \captionsetup{width=.95\linewidth}
    \includegraphics[width=.87\linewidth]{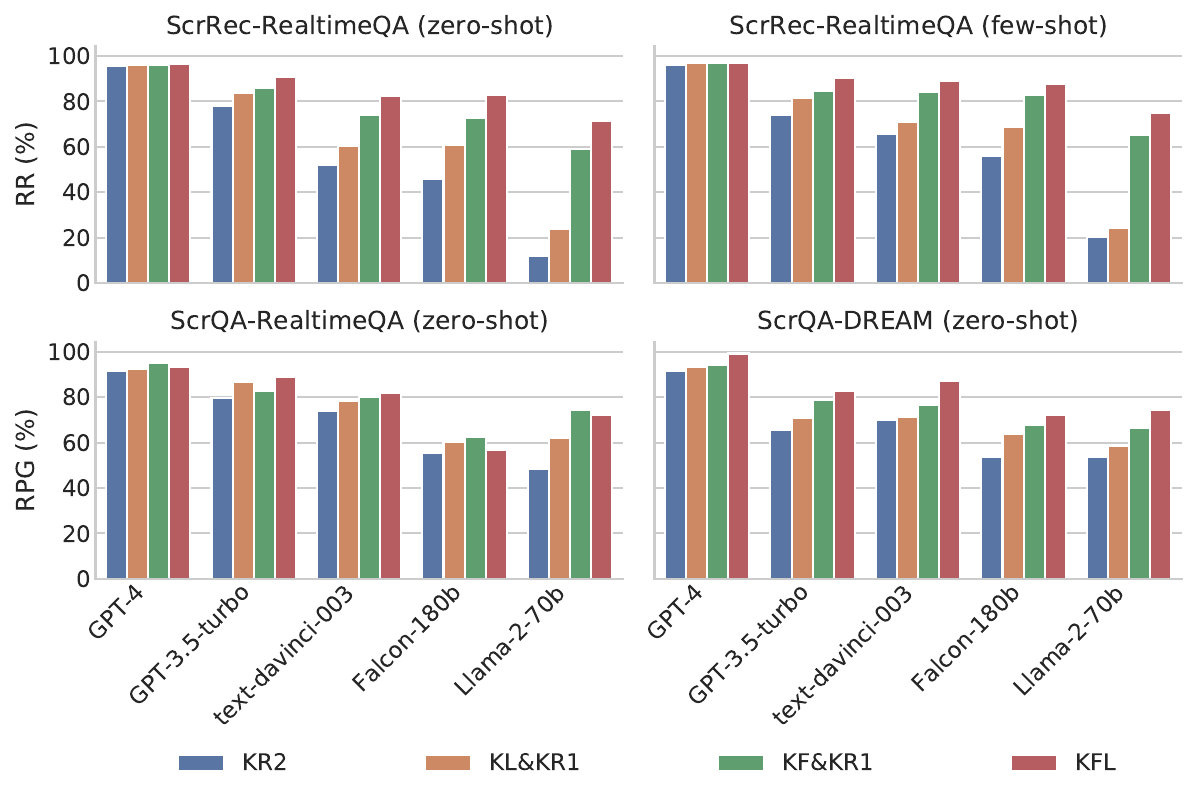}
    \vspace{-0.4cm}
    \caption{Illustrating letter importance by comparing performance of LLMs between four scramble types.}
    \label{fig:importance}
    \vspace{-0.cm}
\end{figure*}

\begin{figure*}[ht!]
    \centering
    \captionsetup{width=.95\linewidth}
    \includegraphics[width=0.99\linewidth]{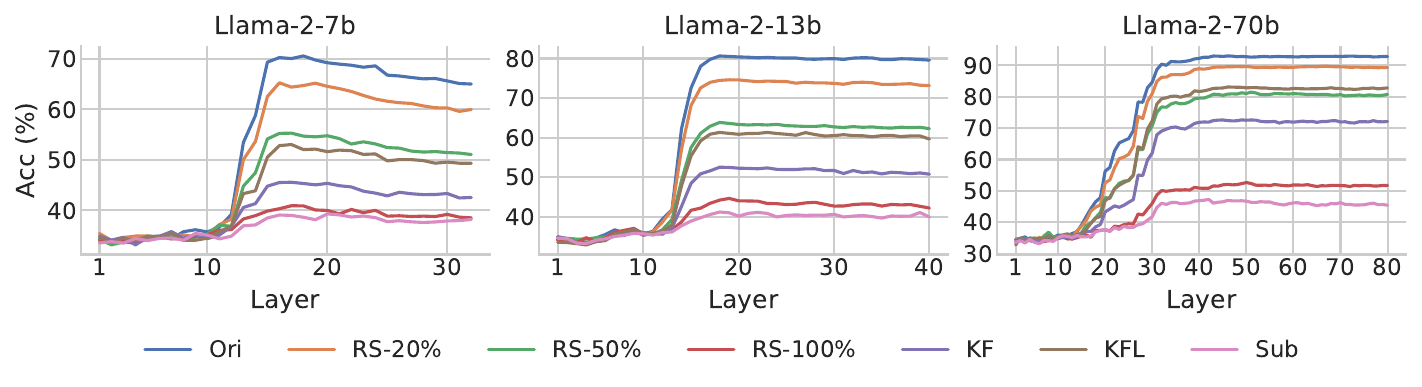}
    \vspace{-0.3cm}
    \caption{Performance of probe classifier using hidden states of each layer to predict answer on DREAM dataset in different scramble settings for three Llama-2 model with different parameter sizes.}
    \label{fig:probe}
    \vspace{-0.1cm}
\end{figure*}
\begin{figure*}[!ht]
    \centering
    \captionsetup{width=.95\linewidth}
    \includegraphics[width=0.999\linewidth]{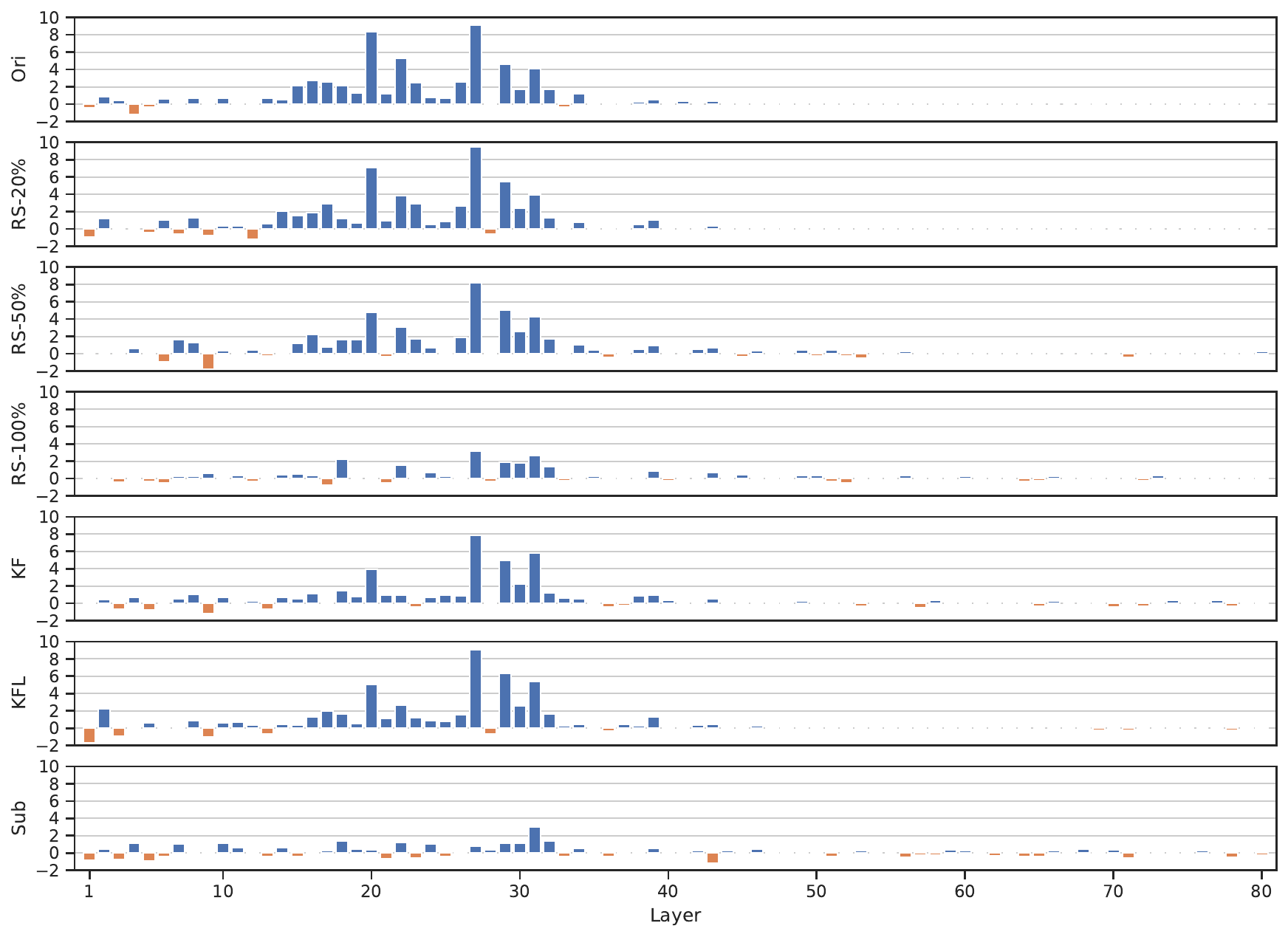}
    \vspace{-0.85cm}
    \caption{Layer-wise accuracy differential: the difference of probe task performance between using hidden states from the current layer and the previous layer for Llama-2-70b in different settings.}
    \label{fig:probe_delta}
    \vspace{-0.1cm}
\end{figure*}
\begin{figure*}[!ht]
    \centering
    \captionsetup{width=.95\linewidth}
    \includegraphics[width=0.999\linewidth]{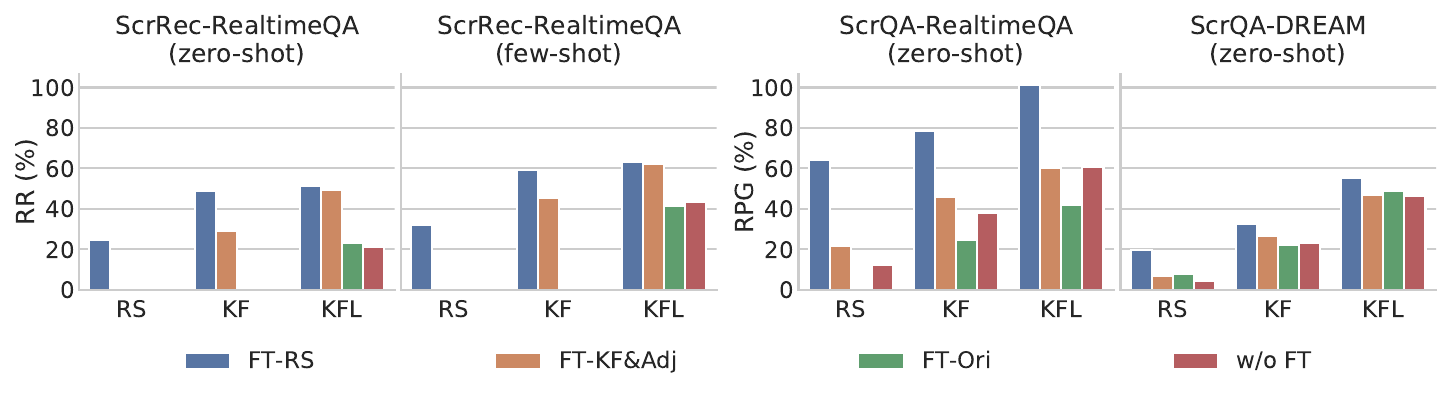}
    \vspace{-0.85cm}
    \caption{Comparison of performance between three finetuned models (FT-RS, FT-KF\&Adj, FT-Ori) and original Llama-2-13b without finetuing (w/o FT) in four task settings. The missing bars represent values less than or equal to zero.}
    \label{fig:train}
    \vspace{-0.1cm}
\end{figure*}
In main experiments, we evaluate LLMs on three scrambled types (i.e., RS, KF and KFL). Nevertheless, the findings do not offer definitive conclusions regarding the varying importance of letter positions in word recognition, as the words to be scrambled vary for each type (e.g., in RS, all words with more than one letter are scrambled, while in KFL, only words with more than three letters are scrambled). To investigate the letter importance in word recognition for LLMs, we add three additional scramble types: (1) KR2, where we keep two letters in random positions unchanged; (2) KL\&KR1, where we keep the last letter and one letter in another random position unchanged; and (3) KF\&KR1, where we keep the first letter and one letter in another random position unchanged. Using these three scramble types, along with KFL, we can mitigate the difference of scrambled words, since same words with more than three letters are scrambled in all four types.

~\autoref{fig:importance} demonstrates the performance of the top five LLMs on four scramble types. The performance of each LLM typically follows an ascending order: KR2, KL\&KR1, KF\&KR1, KFL, with the exception of ScrQA-RealtimeQA. On ScrRec, the gaps between different scramble types are larger for less capable models. However, on ScrQA, the gaps are smaller, especially when using RealtimeQA. This might be the reason why the pattern does not hold on ScrQA-RealtimeQA, as the randomness has a greater impact on performance order when the gaps are smaller. The results imply that the first and last letters are important for LLMs to recognize words, with the first letter being more important than the last letter. This finding is consistent with human behavior~\citep{rawlinson2007significance, mason1982recognition, johnson2012importance}.

\subsection{Probing Hidden States}

To better investigate the internal mechanistic of LLMs to process scrambled text, we design a probe task using the hidden states of Llama-2 models of different parameter sizes (7b, 13b and 70b) when inputting scrambled text~\cite{tenney-etal-2019-bert}. In the experiment, we collect all data in DREAM dataset (including training, development and test sets), and process each sample with various scramble types and rates used in main experiments. The hidden state of last token in each layer (since last token contains all information of each question) is extracted when inputting the question with dialogue in various scramble settings (Ori, RS-20\%, RS-50\%, RS-100\%, KF, KFL, Sub). For each layer, we train a logistic regression model as a probing classifier using a mixture of hidden states from different settings in the training set to predict the answer. We evaluate the performance in different scramble settings separately by using a combination of the development set and test set.

~\autoref{fig:probe} displays the layer-wise prediction accuracy using hidden states from three models. With more layers and more parameters for comprehension, LLMs tend to better comprehend scrambled text with the help of unchanged surrounding context (as in RS-20\%, RS-50\%) and unchanged first and last letters (as in KF, KFL). However, even for Llama-2-70b, comprehending the text in RS-100\% is still too challenging.
~\autoref{fig:probe_delta} shows the layer-wise accuracy differential on probe task for Llama-2-70b in different settings. Ori, RS-20\%, RS-50\%, KF, and KFL exhibit similar patterns in layer distribution, but with varying degrees of change. However, the pattern of RS-100\% is more similar to that of Sub. For example, both RS-100\% and Sub show almost no improvement in the 20th layer, unlike what is observed in other settings.
\begin{table*}[ht!]
    \centering
    \begin{tabular}{l  l  l  l  l  l }
    \hline
    \thead{\textbf{Level}} & \thead{\textbf{Example} \\ (Original: How does the\\ change affect tokenization?) } & \thead{\textbf{Natural} \\ \textbf{errors}} &  \thead{\textbf{Change of} \\ \textbf{tokenization}} & \thead{\textbf{Severity}} & \thead{\textbf{Studies}} \\
    \hline
    \thead{Word} & \thead{How the affect does \\tokenization change?} &\thead{No} &  \thead{Order} & \thead{Low--High} & \thead{~\citealt{sinha-etal-2021-masked},\\~\citealt{sinha-etal-2021-unnatural},\\~\citealt{pham-etal-2021-order},\\~\citealt{abdou-etal-2022-word}} \\
    \hline
    \thead{Sub-word} & \thead{How change token does \\affectization the?} &\thead{No} &  \thead{Order} & \thead{Low--High} & \thead{~\citealt{abdou-etal-2022-word}} \\
    \hline
    \thead{Character} & \thead{How does the chagne\\ affect tokeniztion?} &\thead{Yes} &  \thead{Tokens} & \thead{Low} & \thead{~\citealt{wang2023robustness},\\~\citealt{zhu2023promptbench}} \\
    \hline
    \thead{Character} & \thead{wHo deso teh chenga\\ afctfe aotznoenitki? } &\thead{No} &  \thead{Tokens} & \thead{Middle--Extremely High} & \thead{Ours} \\
    \hline
    \end{tabular}
    \caption{Categorizing the prior studies and demonstrating the position of this work.}
    \label{tab: related_work}
\end{table*}

\subsection{Training on Scrambled Text}

To initially investigate the potential reasons for this phenomenon, we aim is to examine the impact of training on scrambled text. In this experiment, we select Llama-2-13b model as a base model, then finetune it on scrambled text and evaluate finetuned models on scrambled bench. The 20220301.simple subset from wikipedia dataset~\cite{wikidump} is utilized to finetune the model on a language modelling task (irrelevant to our evaluation tasks). We finetune Llama-2-13b on scrambled dataset with two settings: (1) randomly scramble 30\% words with more than two letters in 60\% samples (FT-RS); (2) keep first letter unchanged and swap two adjacent letters in other positions for 30\% words with more than two letters in 60\% samples (FT-KF\&Adj) to simulate natural typographical errors that are more likely to be found in real-world text. Additionally, we finetune a model using the original dataset (FT-Ori) to serve as a control for comparative analysis.

~\autoref{fig:train} demonstrates the comparison of performance between three finetuned models and original Llama-2-13b without finetuing (w/o FT) on various datasets and scramble types. FT-Ori usually gets performance similar to w/o FT  (except for ScrQA-RealtimeQA) which indicates that the original data is not likely to be beneficial for evaluation tasks. FT-RS shows significant improvement over w/o-FT baseline across all tasks, and it achieves positive RR in RS setting, which even surpasses the performance of Falcon-180b. FT-KF\&Adj gets similar or marginally better performance than w/o FT on ScrQA. While, on ScrRec, it improve a lot over w/o FT baseline in KF and KFL, especially it can achieve comparable performance with FT-RS in KFL. It could be inferred that training on text with natural errors is likely to be advantageous for models in handling unnatural scrambled text, especially under KFL condition.

\section{Summary of related work}
\label{sec:app_related_work}
\autoref{tab: related_work} categorizes the related work and demonstrates the position of our study.

\section{Examples of different tasks (by GPT-4)}
\label{sec:examples}
\subsection{Zero-shot ScrRec on RealtimeQA}
\textbf{Input:}\\
The following sentence contains words with scrambled letters. Please recover original sentence from it.\\
Scrambled sentence: A reGebr byba ulfaorm wsa titbudiserd ot soetsr epdstie a lclera eovr bslpioes ionmanantitco, grnoadicc ot eth ADF. heT pyomacn si noniacrggue rptsean ot ckhec yna poducsrt yhte evah ta mhoe nda cdisadr sehot taht aym eb ecaeftdf.\\
Recovered sentence:\\
\textbf{Output:}\\
A Gerber baby formula was distributed to stores despite a recall over possible contamination, according to the FDA. The company is encouraging parents to check any products they have at home and discard those that may be affected.
\subsection{Few-shot ScrRec on RealtimeQA}
\textbf{Input:}\\
Scrambled Sentence: eTh camp continued to fnctinuo this ayw ilntu the rwa needd.\\
Recovered Sentence: The camp continued to function this way until the war ended.\\
\\
Scrambled Sentence: It swa first developed ni the 1980s yb oAcrn Computers tdL ot erowp their pstodke nmecisah and subsequently supn off sa a separate paocnmy, now ARM Holdings.\\
Recovered Sentence: It was first developed in the 1980s by Acorn Computers Ltd to power their desktop machines and subsequently spun off as a separate company, now ARM Holdings.\\
\\
Scrambled Sentence: According to the CIA kcbFotoa, the United States is one fo eethr iusecnort (het etrhos nebgi Liberia nda mBuar/Myanmar) that sha not adopted eth International System fo Utins (SI) rmtcei symset as iethr ffliicao system fo gswheit dna measures.\\
Recovered Sentence: According to the CIA Factbook, the United States is one of three countries (the others being Liberia and Burma/Myanmar) that has not adopted the International System of Units (SI) metric system as their official system of weights and measures.\\
\\
Scrambled sentence: A reGebr byba ulfaorm wsa titbudiserd ot soetsr epdstie a lclera eovr bslpioes ionmanantitco, grnoadicc ot eth ADF. heT pyomacn si noniacrggue rptsean ot ckhec yna poducsrt yhte evah ta mhoe nda cdisadr sehot taht aym eb ecaeftdf.\\
Recovered sentence:\\
\textbf{Output:}\\
A Gerber baby formula was distributed to stores despite a recall over possible contamination, according to the FDA. The company is encouraging parents to check any products they have at home and discard those that may be affected.\\
\subsection{Zero-shot ScrQA on RealtimeQA}
\textbf{Input:}\\
Question: Which type of product was recently distributed to some US stores despite a recall notice?\\
Choices: (A)Salad dressing (B)Baby formula (C)Ground beef (D)Whole milk\\
Evidence: A reGebr byba ulfaorm wsa titbudiserd ot soetsr epdstie a lclera eovr bslpioes ionmanantitco, grnoadicc ot eth ADF. heT pyomacn si noniacrggue rptsean ot ckhec yna poducsrt yhte evah ta mhoe nda cdisadr sehot taht aym eb ecaeftdf.\\
Answer: Based on the evidence, among A through D, the answer is\\
\textbf{Output:}\\
(B)Baby formula
\subsection{Zero-shot ScrQA on DREAM}
\textbf{Input:}\\
Dialogue:\\
M: owH goln veah uyo eben ehctnaig ni iths edmdli colosh?\\
W: Fro ent esray. oT eb akfrn, I'm edirt fo higneact het mesa bokoextt rof os ongl hghuto I od jyneo gebin a aterceh. I'm idenoncisrg gntriy minseghot wne.\\
Question: What's the woman probably going to do?\\
Choices: (A)To teach a different textbook. (B)To change her job. (C)To learn a different textbook.\\
Answer: Based on the dialogue, among A through C, the answer is\\
\textbf{Output:}\\
(B)To change her job.

\subsection{Few-shot CoT ScrQA on AQuA-RAT}
\textbf{(1)}\\
\textbf{Input:}\\
Question: hJno nfodu ttah teh gaeerav fo 15 rnsbmue si 40. fI 10 si edadd ot ceha umenbr enht teh naem fo het rbenmsu si?\\
Choices: (A)50 (B)45 (C)65 (D)78 (E)64\\
Answer: If 10 is added to each number, then the mean of the numbers also increases by 10. So the new mean would be 50. The answer is (A).\\
\\
Question: fI a / b = 3/4 dan 8a + 5b = 22, htne nfid het evual fo a.\\
Choices: (A)1/2 (B)3/2 (C)5/2 (D)4/2 (E)7/2\\
Answer: If a / b = 3/4, then b = 4a / 3. So 8a + 5(4a / 3) = 22. This simplifies to 8a + 20a / 3 = 22, which means 44a / 3 = 22. So a is equal to 3/2. The answer is (B).\\
\\
Question: A nprseo si gtlanrvie ta 20 mk/rh nda hdrecae ihs ynedits ni 2.5 rh neth fdin eth dicatsen?\\
Choices: (A)53 km (B)55 km (C)52 km (D)60 km (E)50 km\\
Answer: The distance that the person traveled would have been 20 km/hr * 2.5 hrs = 50 km. The answer is (E).\\
\\
Question: wHo aymn etosreyksk ear deeend ot peyt het nsrbmeu rfmo 1 ot 500?\\
Choices: (A)1156 (B)1392 (C)1480 (D)1562 (E)1788\\
Answer: There are 9 one-digit numbers from 1 to 9. There are 90 two-digit numbers from 10 to 99. There are 401 three-digit numbers from 100 to 500. 9 + 90(2) + 401(3) = 1392. The answer is (B).\\
\\
Question: A ritna inunrgn ta a eepsd fo 100 mlesi/uhro, eakts 10 hurso ot cerah ist dneitiotsan. trefA ovgeirnc etarqru fo het cntsdaie, ti tstrsa grniina nda eth rniat ash ot eb dlseow ot seedp fo 75 liesm/uhor. Wtha si het otalt orjeyun iroatdnu?\\
Choices: (A)10 (B)11.5 (C)12.5 (D)13.5 (E)15\\
Answer:\\
\textbf{Output:}\\
The train covers a quarter of the distance at 100 miles/hour, so it travels for 2.5 hours (10 hours / 4) at this speed. The remaining distance is 3/4 of the total distance, which takes the train 7.5 hours (10 hours - 2.5 hours) at 100 miles/hour. When the train slows down to 75 miles/hour, it takes 1.5 times longer to cover the same distance (100 miles/hour / 75 miles/hour = 1.33). So the train takes 11.25 hours (7.5 hours * 1.33) to cover the remaining distance at 75 miles/hour. The total journey duration is 13.75 hours (2.5 hours + 11.25 hours). The answer is (D).\\
\\
\textbf{(2)}\\
\textbf{Input:}\\
Question: John found that the average of 15 numbers is 40. If 10 is added to each number then the mean of the numbers is?\\
Choices: (A)50 (B)45 (C)65 (D)78 (E)64\\
Answer: If 10 is added to each number, then the mean of the numbers also increases by 10. So the new mean would be 50. The answer is (A).\\
\\
Question: If a / b = 3/4 and 8a + 5b = 22, then find the value of a.\\
Choices: (A)1/2 (B)3/2 (C)5/2 (D)4/2 (E)7/2\\
Answer: If a / b = 3/4, then b = 4a / 3. So 8a + 5(4a / 3) = 22. This simplifies to 8a + 20a / 3 = 22, which means 44a / 3 = 22. So a is equal to 3/2. The answer is (B).\\
\\
Question: A person is traveling at 20 km/hr and reached his destiny in 2.5 hr then find the distance?\\
Choices: (A)53 km (B)55 km (C)52 km (D)60 km (E)50 km\\
Answer: The distance that the person traveled would have been 20 km/hr * 2.5 hrs = 50 km. The answer is (E).\\
\\
Question: How many keystrokes are needed to type the numbers from 1 to 500?\\
Choices: (A)1156 (B)1392 (C)1480 (D)1562 (E)1788\\
Answer: There are 9 one-digit numbers from 1 to 9. There are 90 two-digit numbers from 10 to 99. There are 401 three-digit numbers from 100 to 500. 9 + 90(2) + 401(3) = 1392. The answer is (B).\\
\\
Question: ieahMnc A psut otu a oy-oy ryeve 6 nsimtue. enicMah B pstu uto a oy-oy veeyr 9 nmusite. tAefr hwo aynm smtnuei lwil yeth hvea eurdpcod 10 oy-soy?\\
Choices: (A)24 minutes (B)32 minutes (C)36 minutes (D)64 minutes (E)72 minutes\\
Answer:\\
\textbf{Output:}\\
First, let's rewrite the question with the words unscrambled: "Machine A puts out a yo-yo every 6 minutes. Machine B puts out a yo-yo every 9 minutes. After how many minutes will they have produced 10 yo-yos?"\\
\\
To solve this problem, we can find the least common multiple (LCM) of 6 and 9, which is 18. This means that every 18 minutes, both machines will have produced a yo-yo at the same time. In 18 minutes, Machine A will have produced 3 yo-yos (18/6) and Machine B will have produced 2 yo-yos (18/9), for a total of 5 yo-yos. Since we need 10 yo-yos, we can multiply 18 minutes by 2 to get 36 minutes. The answer is (C) 36 minutes.\\

\section{Full experimental results}
\label{sec:full_results}
We conduct experiments using the most powerful closed-source LLMs, including text-davinci-003, GPT-3.5-turbo and GPT-4 and various open-source LLMs, including the models from Falcon series, Llama-2 series, MPT series, UL2 series, and T5 series. The open-source model covers diverse model architectures (decoder only and encoder-decoder), model size (from 7b to 180b), training objectives (e.g., with or without further finetuning) and tokenizers (e.g., tokenizer-free: ByT5-xxl).

For GPT-4, the version GPT-4-0314 is used. For GPT-3.5-turbo, the version GPT-3.5-turbo-0301 is used. For Falcon-180b and Falcon-180b-chat, the quantized method~\cite{dettmers2023qlora} is used to load the model and run the experiments. It probably affects their performance to some extent.

\subsection{Full experimental results on scrambled RealtimeQA dataset}

~\autoref{tab:realtime_rec_few} and~\autoref{tab:realtime_rec_zero} illustrate the full experimental results on ScrRec using the scrambled RealtimeQA dataset in both few-shot and zero-shot settings, respectively.

~\autoref{tab:realtime_qa} illustrates the full experimental results on ScrQA using scrambled RealtimeQA dataset in a zero-shot setting.

\subsection{Full experimental results on scrambled DREAM dataset}

~\autoref{tab:dream_qa_all} illustrate the full experimental results on ScrQA using scrambled DREAM dataset in a zero-shot setting.

~\autoref{tab:dream_qa_cat_acc} and~\autoref{tab:dream_qa_cat_rpg}
display the experimental results for various question categories, comparing the top five models using Acc and RPG as evaluation metrics.
\begin{table*}
    \centering
    \begin{tabular}{cccccl} 
    \hline
             \textbf{Model} & \textbf{20\%} & \textbf{50\%} & \textbf{100\%} & \textbf{KF} & \textbf{KFL}\\
             \hline
 \multirow{2}{*}{GPT-4} & 1.49 & 2.40 & 4.79 &3.79 &2.24\\ 
                            & 94.51 & 96.46 & 96.45 &96.37 &97.04\\ \cline{1-6}
 
 \multirow{2}{*}{GPT-3.5-turbo} &  5.80 &  15.38 &  36.87 & 20.76 &7.40\\ 
                                    &  78.57 &  77.29 &  72.63 & 80.14 &90.22\\ \cline{1-6}
 \multirow{2}{*}{text-davinci-003} &  5.38 &  18.89 &  84.39 & 22.21 &8.36\\ 
                           &  80.13 &  72.11 &  37.35 & 78.75 &88.95\\ \cline{1-6}
 \multirow{2}{*}{Falcon-180b} &  4.18 &  21.84 &  116.73 & 28.77 &9.40\\ 
                           &  84.55 &  67.76 &  13.34 & 72.47 &87.58\\ \cline{1-6}
 \multirow{2}{*}{Falcon-180b-chat} &  11.61 &  38.19 &  186.65 & 63.51 &21.73\\ 
                           &  57.12 &  43.62 &  < 0 & 39.24 &71.28\\ \cline{1-6}
 \multirow{2}{*}{Falcon-40b} &  17.27 &  69.86 &  185.26 & 119.00 &66.03\\ 
                           &  36.18 &  < 0 &  < 0 & < 0 &12.71\\ \cline{1-6}
 \multirow{2}{*}{Falcon-40b-instruct }&  18.79 &  74.26 &  192.86 & 130.82 &62.39\\ 
                           &  30.56 &  < 0 &  < 0 & < 0 &17.53\\ \cline{1-6}
 \multirow{2}{*}{Llama-2-70b} &  7.85 &  37.30 &  169.85 & 56.25 &18.99\\ 
                            &  71.0 &  44.93 &  < 0 & 46.19 &74.9\\ \cline{1-6}
 \multirow{2}{*}{Llama-2-70b-chat} &  11.11 &  54.49 &  196.44 & 106.31 &43.18\\ 
                           &  58.97 &  19.56 &  < 0 & < 0 &42.93\\ \cline{1-6}
 \multirow{2}{*}{Llama-2-13b} &  14.18 &  59.13 &  192.67 & 103.50 &42.67\\ 
                            &  47.62 &  12.71 &  < 0 & 0.99 &43.59\\ \cline{1-6}
 \multirow{2}{*}{Llama-2-13b-chat} & 20.75 & 73.04 & 201.50 &119.52 &56.03\\ 
                            & 23.35 & < 0 & < 0 &< 0 &25.93\\ \cline{1-6}
 \multirow{2}{*}{Llama-2-7b} & 20.90 & 79.83 & 194.48 &135.50 &68.18\\ 
                               & 22.77 & < 0 & < 0 &< 0 &9.88\\\cline{1-6}
 \multirow{2}{*}{Llama-2-7b-chat} & 25.14 & 83.13 & 196.88 &132.44 &73.84\\ 
                                     & 7.11 & < 0 & < 0 &< 0 &2.39\\ \cline{1-6}
 \multirow{2}{*}{MPT-30b} & 14.36 & 54.81 & 181.43 &90.92 &36.14\\ 
                           & 46.96 & 19.08 & < 0 &13.02 &52.23\\ \cline{1-6}
 \multirow{2}{*}{MPT-30b-instruct} & 19.36 & 66.05 & 170.88 &99.38 &50.44\\ 
                                  & 28.46 & 2.49 & < 0 &4.92 &33.33\\   \cline{1-6}
 \multirow{2}{*}{UL2} & 18.12 & 68.79 & 153.53 &109.47 &71.70\\ 
                       & 33.03 & < 0 & < 0 &< 0 &5.22\\     \cline{1-6}
 \multirow{2}{*}{Flan-UL2} & 20.51 & 62.91 & 137.15 &105.67 &74.35\\ 
                   & 24.21 & 7.13 & < 0 &< 0 &1.72\\       \cline{1-6}   
 \multirow{2}{*}{Flan-T5-xxl} & 16.60 & 59.34 & 136.41 &105.44 &73.97\\ 
                           & 38.68 & 12.4 & < 0 &< 0 &2.22\\ \cline{1-6}
 \multirow{2}{*}{ByT5-xxl} & 79.27 & 104.28 & 160.17 &123.97 &102.50\\ 
                           & < 0 & < 0 & < 0 &< 0 &< 0\\ \cline{1-6}
 \hline
    \end{tabular}
    \caption{The experimental results on few-shot ScrRec using RealtimeQA dataset (upper: ED; lower: RR).}
    \label{tab:realtime_rec_few}
\end{table*}

\begin{table*}
    \centering
    \begin{tabular}{cccccl} 
    \hline
             \textbf{Model} & \textbf{20\%} & \textbf{50\%} & \textbf{100\%} & \textbf{KF} & \textbf{KFL}\\
             \hline
 \multirow{2}{*}{GPT-4} & 1.69 & 2.78 & 5.47 &4.19 &2.53\\ 
                             & 93.76 & 95.89 & 95.94 &95.99 &96.66\\  \cline{1-6}
 \multirow{2}{*}{GPT-3.5-turbo} &  4.68 &  13.00 &  29.80 & 14.22 &6.94\\ 
                                     &  82.72 &  80.82 &  77.88 & 86.39 &90.82\\  \cline{1-6}
 \multirow{2}{*}{text-davinci-003} &  15.02 &  31.86 &  109.63 & 30.88 &13.30\\ 
                                   &  44.52 &  52.96 &  18.61 & 70.45 &82.41\\  \cline{1-6}
 \multirow{2}{*}{Falcon-180b} &  7.18 &  26.53 &  159.36 & 43.63 &13.08\\ 
                              &  73.48 &  60.84 &  < 0 & 58.25 &82.72\\  \cline{1-6}
 \multirow{2}{*}{Falcon-180b-chat} &  15.92 &  50.50 &  180.13 & 76.31 &29.29\\ 
                                   &  41.17 &  25.46 &  < 0 & 27.0 &61.29\\  \cline{1-6}
 \multirow{2}{*}{Falcon-40b} &  20.59 &  66.33 &  177.33 & 92.45 &39.04\\ 
                             &  23.94 &  2.08 &  < 0 & 11.56 &48.39\\  \cline{1-6}
 \multirow{2}{*}{Falcon-40b-instruct} &  26.42 &  78.94 &  178.55 & 104.46 &48.42\\ 
                                      &  2.4 &  < 0 &  < 0 & 0.07 &36.0\\  \cline{1-6}
 \multirow{2}{*}{Llama-2-70b} &  8.11 &  41.10 &  180.82 & 60.16 &21.68\\ 
                                 &  70.04 &  39.32 &  < 0 & 42.45 &71.34\\  \cline{1-6}
 \multirow{2}{*}{Llama-2-70b-chat} &  12.65 &  60.23 &  181.30 & 118.13 &51.73\\ 
                                      &  53.26 &  11.08 &  < 0 & < 0 &31.62\\  \cline{1-6}
 \multirow{2}{*}{Llama-2-13b} &  20.21 &  75.54 &  231.41 & 127.63 &59.77\\ 
                                 &  25.32 &  < 0 &  < 0 & < 0 &20.99\\  \cline{1-6}
 \multirow{2}{*}{Llama-2-13b-chat} & 26.79 & 84.09 & 191.85 &120.51 &63.82\\ 
                                      & 1.03 & < 0 & < 0 &< 0 &15.64\\  \cline{1-6}
 \multirow{2}{*}{Llama-2-7b} & 29.81 & 106.38 & 192.05 &165.56 &88.65\\ 
                                & < 0 & < 0 & < 0 &< 0 &< 0\\  \cline{1-6}
 \multirow{2}{*}{Llama-2-7b-chat} & 48.92 & 123.66 & 202.16 &167.49 &112.03\\ 
                                     & < 0 & < 0 & < 0 &< 0 &< 0\\  \cline{1-6}
 \multirow{2}{*}{MPT-30b} & 20.54 & 70.69 & 226.34 &135.58 &62.16\\ 
                          & 24.11 & < 0 & < 0 &< 0 &17.84\\  \cline{1-6}
 \multirow{2}{*}{MPT-30b-instruct} & 36.01 & 84.98 & 233.60 &132.77 &76.60\\ 
                                   & < 0 & < 0 & < 0 &< 0 &< 0\\  \cline{1-6}
 \multirow{2}{*}{UL2} & 72.39 & 139.14 & 205.86 &188.57 &163.67\\ 
                     & < 0 & < 0 & < 0 &< 0 &< 0\\  \cline{1-6}
 \multirow{2}{*}{Flan-UL2} & 33.10 & 71.76 & 136.94 &106.48 &78.55\\ 
                           & < 0 & < 0 & < 0 &< 0 &< 0\\  \cline{1-6}
 \multirow{2}{*}{Flan-T5-xxl} & 27.07 & 68.02 & 136.47 &106.08 &76.69\\ 
                              & < 0 & < 0 & < 0 &< 0 &< 0\\  \cline{1-6}
 \multirow{2}{*}{ByT5-xxl} & 85.83 & 111.53 & 155.27 &136.99 &117.80\\ 
                           & < 0 & < 0 & < 0 &< 0 &< 0\\  \cline{1-6}
 \hline
    \end{tabular}
    \caption{The experimental results on zero-shot ScrRec using RealtimeQA dataset (upper: ED; lower: RR).}
    \label{tab:realtime_rec_zero}
\end{table*}

\begin{table*}
    \centering
    \begin{tabular}{cccccccl} 
    \hline
             \textbf{Model} & \textbf{Ori} & \textbf{20\%} & \textbf{50\%} & \textbf{100\%} & \textbf{KF} & \textbf{KFL}  & \textbf{Sub}\\
             \hline
\multirow{2}{*}{GPT-4}  & 97.69 & 97.40 & 96.82 & 93.35 & 93.35 & 95.66 & 62.14\\
                            & 100.0 & 99.19 & 97.56 & 87.80 & 87.80 & 94.31 & 0.0\\  \cline{1-8}
\multirow{2}{*}{GPT-3.5-turbo} & 97.11 & 94.80 & 91.33 & 81.79 & 88.44 & 93.06 & 52.60 \\
                                    & 100.0 & 94.81 & 87.01 & 65.58 & 80.52 & 90.91 & 0.0\\ \cline{1-8}
\multirow{2}{*}{text-davinci-003} & 93.93 & 91.62 & 87.28 & 77.46 & 84.10 & 88.15 & 54.91\\
                                  & 100.0 & 94.07 & 82.96 & 57.78 & 74.81 & 85.19 & 0.0\\ \cline{1-8}
\multirow{2}{*}{Falcon-180b} & 84.68 & 82.37 & 78.61 & 65.03 & 70.52 & 74.28  & 57.23\\
                              & 100.0 & 91.58 & 77.89 & 28.42 & 48.42 & 62.11  & 0.0 \\ \cline{1-8}
\multirow{2}{*}{Falcon-180b-chat}  & 75.43 & 72.54 & 71.39 & 58.09 & 63.87 & 67.34  & 54.62\\ 
                                  & 100.0 & 86.11 & 80.56 & 16.67 & 44.44 & 61.11 & 0.0\\ \cline{1-8}
\multirow{2}{*}{Falcon-40b} & 58.96 & 59.25 & 54.62 & 44.51 & 44.51 & 50.29 & 39.60 \\
                           & 100.0 & 101.49 & 77.61 & 25.37 & 25.37 & 55.22 & 0.0 \\ \cline{1-8}
\multirow{2}{*}{Falcon-40b-instruct} & 65.90 & 63.87 & 61.27 & 48.55 & 53.47 & 60.69 & 46.82\\
                                     & 100.0 & 89.39 & 75.76 & 9.09 & 34.85 & 72.73 & 0.0\\ \cline{1-8}
\multirow{2}{*}{Llama-2-70b}& 88.44 & 86.13 & 80.35 & 67.92 & 75.43 & 80.64  & 57.23\\
                            & 100.0 & 92.59 & 74.07 & 34.26 & 58.33 & 75.00 & 0.0\\ \cline{1-8}
\multirow{2}{*}{Llama-2-70b-chat}& 96.24 & 92.49 & 83.82 & 64.74 & 75.43 & 82.66 & 54.91\\ 
                                 & 100.0 & 90.91 & 69.93 & 23.78 & 49.65 & 67.13 & 0.0 \\ \cline{1-8}
\multirow{2}{*}{Llama-2-13b}& 66.18 & 65.03 & 57.51 & 47.40 & 52.89 & 57.80 & 44.80 \\
                            & 100.0 & 94.59 & 59.46 & 12.16 & 37.84 & 60.81 & 0.0 \\ \cline{1-8}
\multirow{2}{*}{Llama-2-13b-chat}& 84.68 & 78.90 & 66.18 & 47.98 & 55.20 & 59.83 & 44.51 \\
                                & 100.0 & 85.61 & 53.96 & 8.63 & 26.62 & 38.13 & 0.0\\ \cline{1-8}
\multirow{2}{*}{Llama-2-7b}& 47.40 & 45.09 & 42.77 & 38.73 & 40.75 & 43.64 & 39.31 \\
                          & 100.0 & 71.43 & 42.86 & < 0 & 17.86 & 53.57 & 0.0 \\ \cline{1-8}
\multirow{2}{*}{Llama-2-7b-chat} & 82.08 & 75.14 & 64.16 & 45.09 & 52.31 & 56.94 & 43.93 \\
                                & 100.0 & 81.82 & 53.03 & 3.03 & 21.97 & 34.09 & 0.0 \\ \cline{1-8}
\multirow{2}{*}{MPT-30b}& 62.72 & 56.36 & 50.87 & 37.86 & 47.69 & 53.47 & 37.28 \\
                         & 100.0 & 75.00 & 53.41 & 2.27 & 40.91 & 63.64 & 0.0 \\ \cline{1-8}
\multirow{2}{*}{MPT-30b-instruct} & 65.03 & 66.76 & 58.67 & 47.11 & 54.62 & 61.27 & 39.88 \\
                                  & 100.0 & 106.90 & 74.71 & 28.74 & 58.62 & 85.06 & 0.0 \\ \cline{1-8}
\multirow{2}{*}{UL2} & 87.57 & 84.68 & 76.30 & 45.95 & 60.69 & 71.10 & 38.44 \\
                     & 100.0 & 94.12 & 77.06 & 15.29 & 45.29 & 66.47 & 0.0 \\     \cline{1-8}
\multirow{2}{*}{Flan-UL2}  & 98.55 & 95.95 & 89.31 & 49.71 & 65.61 & 76.88  & 46.24\\
                          & 100.0 & 95.03 & 82.32 & 6.63 & 37.02 & 58.56 & 0.0 \\ \cline{1-8}
\multirow{2}{*}{Flan-T5-xxl} & 97.40 & 96.24 & 87.86 & 49.42 & 66.47 & 82.66 & 40.75 \\
                             & 100.0 & 97.96 & 83.16 & 15.31 & 45.41 & 73.98 & 0.0 \\ \cline{1-8}
\multirow{2}{*}{ByT5-xxl} & 73.12 & 57.23 & 46.24 & 26.30 & 35.84 & 41.91 & 25.14 \\
                          & 100.0 & 66.87 & 43.98 & 2.41 & 22.29 & 34.94  & 0.0\\ \cline{1-8}

 \hline
    \end{tabular}
    \caption{The experimental results on zero-shot ScrQA using RealtimeQA dataset (upper: Acc; lower: RPG).}
    \label{tab:realtime_qa}
\end{table*}

\begin{table*}
    \centering
    \begin{tabular}{cccccccl} 
    \hline
             \textbf{Model} & \textbf{Ori} & \textbf{20\%} & \textbf{50\%} & \textbf{100\%}  & \textbf{KF} & \textbf{KFL} & \textbf{Sub}\\
             \hline
 \multirow{2}{*}{GPT-4}  & 96.68 & 96.29 & 94.43 & 87.40 & 93.65 & 96.39 & 46.48 \\
                             & 100.0 & 99.22 & 95.53 & 81.52 & 93.97 & 99.42 & 0.0 \\ \cline{1-8}
 \multirow{2}{*}{GPT-3.5-turbo} & 92.48 & 87.30 & 77.83 & 45.61 & 71.78 & 83.89  & 26.07\\
                                     & 100.0 & 92.21 & 77.94 & 29.41 & 68.82 & 87.06 & 0.0 \\ \cline{1-8}
 \multirow{2}{*}{text-davinci-003} & 92.29 & 89.45 & 80.27 & 52.44 & 76.66 & 87.50 & 43.85 \\
                                   & 100.0 & 94.15 & 75.20 & 17.74 & 67.74 & 90.12 & 0.0 \\  \cline{1-8}
 \multirow{2}{*}{Falcon-180b} & 93.65 & 89.65 & 82.23 & 55.37 & 75.39 & 84.47 & 51.17 \\
                              & 100.0 & 90.57 & 73.10 & 9.89 & 57.01 & 78.39 & 0.0 \\ \cline{1-8}
 \multirow{2}{*}{Falcon-180b-chat} & 92.68 & 89.16 & 78.32 & 49.90 & 70.51 & 80.76 & 45.31 \\
                                   & 100.0 & 92.58 & 69.69 & 9.69 & 53.20 & 74.85 & 0.0 \\ \cline{1-8}
 \multirow{2}{*}{Falcon-40b}  & 70.12 & 65.04 & 56.45 & 44.34 & 50.78 & 58.30 & 43.26 \\ 
                              & 100.0 & 81.09 & 49.09 & 4.00 & 28.00 & 56.00 & 0.0 \\ \cline{1-8}
 \multirow{2}{*}{Falcon-40b-instruct} & 75.20 & 67.48 & 58.59 & 43.65 & 50.00 & 60.06 & 42.87\\
                                      & 100.0 & 76.13 & 48.64 & 2.42 & 22.05 & 53.17 & 0.0 \\ \cline{1-8}
 \multirow{2}{*}{Llama-2-70b} & 93.26 & 89.75 & 80.96 & 54.30 & 73.34 & 84.38 & 47.85\\
                                 & 100.0 & 92.26 & 72.90 & 14.19 & 56.13 & 80.43 & 0.0 \\ \cline{1-8}
 \multirow{2}{*}{Llama-2-70b-chat} & 91.11 & 85.74 & 75.10 & 48.24 & 59.67 & 74.02 & 45.12 \\
                                  & 100.0 & 88.32 & 65.18 & 6.79 & 31.63 & 62.85 & 0.0 \\ \cline{1-8}
 \multirow{2}{*}{Llama-2-13b} & 69.53 & 63.87 & 53.32 & 38.67 & 44.63 & 52.25 & 37.21 \\
                                 & 100.0 & 82.48 & 49.85 & 4.53 & 22.96 & 46.53 & 0.0 \\ \cline{1-8}
 \multirow{2}{*}{Llama-2-13b-chat} & 78.61 & 71.78 & 59.77 & 41.80 & 48.44 & 57.42 & 40.92 \\
                                      & 100.0 & 81.87 & 50.00 & 2.33 & 19.95 & 43.78 & 0.0 \\ \cline{1-8}
 \multirow{2}{*}{Llama-2-7b} & 60.55 & 55.08 & 47.46 & 37.60 & 43.07 & 46.39 & 37.30 \\
                                & 100.0 & 76.47 & 43.70 & 1.26 & 24.79 & 39.08 & 0.0 \\ \cline{1-8}
 \multirow{2}{*}{Llama-2-7b-chat}  & 73.63 & 64.45 & 56.25 & 40.04 & 45.41 & 53.81 & 39.84 \\
                                     & 100.0 & 72.83 & 48.55 & 0.58 & 16.47 & 41.33 & 0.0 \\ \cline{1-8}
 \multirow{2}{*}{MPT-30b} & 74.02 & 69.24 & 59.47 & 47.17 & 50.00 & 57.71 & 43.46 \\
                          & 100.0 & 84.35 & 52.40 & 12.14 & 21.41 & 46.65 & 0.0 \\ \cline{1-8}
 \multirow{2}{*}{MPT-30b-instruct} & 83.50 & 75.68 & 65.43 & 44.82 & 54.30 & 60.64 & 41.41 \\
                                   & 100.0 & 81.44 & 57.08 & 8.12 & 30.63 & 45.71 & 0.0 \\ \cline{1-8}
 \multirow{2}{*}{UL2}  & 43.36 & 39.36 & 29.88 & 14.55 & 17.38 & 25.20 & 12.79 \\
                      & 100.0 & 86.90 & 55.91 & 5.75 & 15.02 & 40.58 & 0.0 \\ \cline{1-8}
 \multirow{2}{*}{Flan-UL2} & 94.24 & 91.21 & 80.96 & 49.80 & 62.99 & 75.39 & 48.14 \\
                           & 100.0 & 93.43 & 71.19 & 3.60 & 32.20 & 59.11 & 0.0 \\ \cline{1-8}
 \multirow{2}{*}{Flan-T5-xxl} & 95.02 & 92.87 & 84.67 & 51.17 & 67.09 & 81.54 & 51.07 \\
                              & 100.0 & 95.11 & 76.44 & 0.22 & 36.44 & 69.33 & 0.0 \\ \cline{1-8}
 \multirow{2}{*}{ByT5-xxl} & 43.16 & 40.23 & 36.23 & 31.93 & 33.50 & 36.62 & 31.84 \\
                             & 100.0 & 74.14 & 38.79 & 0.86 & 14.66 & 42.24 & 0.0 \\ \cline{1-8}
 \hline
    \end{tabular}
    \caption{The experimental results on zero-shot ScrQA using DREAM dataset (upper: Acc; lower: RPG).}
    \label{tab:dream_qa_all}
\end{table*}

\begin{table*}
    \centering
    \begin{tabular}{clcccccll} 
    \hline
             \textbf{Model}  &\textbf{Type}& \textbf{Ori} & \textbf{20\%} & \textbf{50\%} & \textbf{100\%} & \textbf{KF} & \textbf{KFL} &\textbf{Sub} \\
             \hline
{GPT-4} & all & 96.68 & 96.29 & 94.43 & 87.40 & 93.65 & 96.39  &46.48 \\
 & arithmetic & 76.09 & 78.26 & 67.39 & 45.65 & 58.70 & 71.74  &32.61 \\
 & commonsense & 94.69 & 94.41 & 93.02 & 82.40 & 91.90 & 94.41  &42.46 \\
 & logic & 95.97 & 95.84 & 93.42 & 85.77 & 92.62 & 95.97  &47.65 \\
 & matching & 99.07 & 97.22 & 98.15 & 92.59 & 97.22 & 99.07  &47.22 \\
 & summary & 98.47 & 97.71 & 96.18 & 90.08 & 95.42 & 96.95  &36.64 \\  \hline
\multirow{6}{*}{GPT-3.5-turbo} & all & 92.48 & 87.30 & 77.83 & 45.61 & 71.78 & 83.89  &26.07 \\
 & arithmetic & 60.87 & 60.87 & 52.17 & 21.74 & 41.30 & 50.00  &15.22 \\
 & commonsense & 89.11 & 84.36 & 76.54 & 39.39 & 63.69 & 79.61  &25.42 \\
 & logic & 91.01 & 85.64 & 76.11 & 44.83 & 71.28 & 82.95  &25.10 \\
 & matching & 99.07 & 91.67 & 81.48 & 56.48 & 78.70 & 89.81  &30.56 \\
 & summary & 95.42 & 93.13 & 82.44 & 41.22 & 70.99 & 83.97  &24.43 \\ \hline
\multirow{6}{*}{text-davinci-003} & all & 92.29 & 89.45 & 80.27 & 52.44 & 76.66 & 87.50  &43.85 \\
 & arithmetic & 47.83 & 45.65 & 41.30 & 21.74 & 41.30 & 45.65  &17.39 \\
 & commonsense & 90.78 & 87.99 & 77.37 & 46.37 & 71.79 & 84.92  &45.25 \\
 & logic & 91.01 & 87.92 & 78.39 & 52.89 & 75.03 & 86.71  &41.34 \\
 & matching & 99.07 & 94.44 & 86.11 & 57.41 & 87.96 & 93.52  &48.15 \\
 & summary & 94.66 & 94.66 & 86.26 & 48.09 & 77.86 & 87.79  &49.62 \\ \hline
\multirow{6}{*}{Falcon-180b} & all & 93.65 & 89.65 & 82.23 & 55.37 & 75.39 & 84.47  &51.17 \\
 & arithmetic & 52.17 & 47.83 & 45.65 & 34.78 & 43.48 & 47.83  &34.78 \\
 & commonsense & 91.34 & 88.55 & 79.33 & 56.15 & 74.58 & 83.24  &52.23 \\
 & logic & 92.62 & 88.59 & 79.87 & 52.35 & 73.02 & 82.55  &50.47 \\
 & matching & 98.15 & 90.74 & 87.04 & 62.96 & 78.70 & 89.81  &51.85 \\
 & summary & 96.18 & 93.89 & 89.31 & 67.18 & 87.02 & 91.60  &56.49 \\ \hline
\multirow{6}{*}{Llama-2-70b}& all & 93.26 & 89.75 & 80.96 & 54.30 & 73.34 & 84.38 &47.85 \\
 & arithmetic & 58.70 & 47.83 & 50.00 & 36.96 & 47.83 & 43.48 &36.96 \\
 & commonsense & 91.34 & 89.39 & 80.73 & 53.63 & 73.18 & 84.36 &44.97 \\
 & logic & 92.62 & 88.99 & 79.73 & 54.50 & 73.02 & 84.16 &48.05 \\
 & matching & 94.44 & 89.81 & 79.63 & 50.93 & 73.15 & 79.63 &50.93 \\
 & summary & 96.18 & 94.66 & 86.26 & 58.02 & 79.39 & 90.84 &50.38 \\

 \hline
    \end{tabular}
    \caption{The experimental results of different categories on zero-shot ScrQA using DREAM dataset (Acc).}
    \label{tab:dream_qa_cat_acc}
\end{table*}
\begin{table*}
    \centering
 \begin{tabular}{clcccccll} 
    \hline
             \textbf{Model}  &\textbf{Type}& \textbf{Ori} & \textbf{20\%} & \textbf{50\%} & \textbf{100\%} & \textbf{KF} & \textbf{KFL} &\textbf{Sub}\\
          
             \hline
{GPT-4} & all & 100.0 & 99.22 & 95.53 & 81.52 & 93.97 & 99.42  &0.0 \\
 & arithmetic & 100.0 & 105.00 & 80.00 & 30.00 & 60.00 & 90.00  &0.0 \\
 & commonsense & 100.0 & 99.47 & 96.79 & 76.47 & 94.65 & 99.47  &0.0 \\
 & logic & 100.0 & 99.72 & 94.72 & 78.89 & 93.06 & 100.00  &0.0 \\
 & matching & 100.0 & 96.43 & 98.21 & 87.50 & 96.43 & 100.00  &0.0 \\
 & summary & 100.0 & 98.77 & 96.30 & 86.42 & 95.06 & 97.53  &0.0 \\  \hline
\multirow{6}{*}{GPT-3.5-turbo} & all & 100.0 & 92.21 & 77.94 & 29.41 & 68.82 & 87.06  &0.0 \\
 & arithmetic & 100.0 & 100.00 & 80.95 & 14.29 & 57.14 & 76.19  &0.0 \\
 & commonsense & 100.0 & 92.54 & 80.26 & 21.93 & 60.09 & 85.09  &0.0 \\
 & logic & 100.0 & 91.85 & 77.39 & 29.94 & 70.06 & 87.78  &0.0 \\
 & matching & 100.0 & 89.19 & 74.32 & 37.84 & 70.27 & 86.49  &0.0 \\
 & summary & 100.0 & 96.77 & 81.72 & 23.66 & 65.59 & 83.87  &0.0 \\
\multirow{6}{*}{text-davinci-003} & all & 100.0 & 94.15 & 75.20 & 17.74 & 67.74 & 90.12  &0.0 \\  \hline
 & arithmetic & 100.0 & 92.86 & 78.57 & 14.29 & 78.57 & 92.86  &0.0 \\
 & commonsense & 100.0 & 93.87 & 70.55 & 2.45 & 58.28 & 87.12  &0.0 \\
 & logic & 100.0 & 93.78 & 74.59 & 23.24 & 67.84 & 91.35  &0.0 \\
 & matching & 100.0 & 90.91 & 74.55 & 18.18 & 78.18 & 89.09  &0.0 \\
 & summary & 100.0 & 100.00 & 81.36 & < 0 & 62.71 & 84.75  &0.0 \\  \hline
\multirow{6}{*}{Falcon-180b} & all & 100.0 & 90.57 & 73.10 & 9.89 & 57.01 & 78.39  &0.0 \\
 & arithmetic & 100.0 & 75.00 & 62.50 & 0.00 & 50.00 & 75.00  &0.0 \\
 & commonsense & 100.0 & 92.86 & 69.29 & 10.00 & 57.14 & 79.29  &0.0 \\
 & logic & 100.0 & 90.45 & 69.75 & 4.46 & 53.50 & 76.11  &0.0 \\
 & matching & 100.0 & 84.00 & 76.00 & 24.00 & 58.00 & 82.00  &0.0 \\
 & summary & 100.0 & 94.23 & 82.69 & 26.92 & 76.92 & 88.46  &0.0 \\  \hline
\multirow{6}{*}{Llama-2-70b}& all & 100.0 & 92.26 & 72.90 & 14.19 & 56.13 & 80.43 &0.0 \\
 & arithmetic & 100.0 & 50.00 & 60.00 & 0.00 & 50.00 & 30.00 &0.0 \\
 & commonsense & 100.0 & 95.78 & 77.11 & 18.67 & 60.84 & 84.94 &0.0 \\
 & logic & 100.0 & 91.87 & 71.08 & 14.46 & 56.02 & 81.02 &0.0 \\
 & matching & 100.0 & 89.36 & 65.96 & 0.00 & 51.06 & 65.96 &0.0 \\
 & summary & 100.0 & 96.67 & 78.33 & 16.67 & 63.33 & 88.33 &0.0 \\  \hline
    \end{tabular}
    \caption{The experimental results of different categories on zero-shot ScrQA using DREAM dataset (RPG).}
    \label{tab:dream_qa_cat_rpg}
\end{table*}

\end{document}